\documentclass[final,3p]{elsarticle}

\usepackage{lineno}
\usepackage[colorlinks=true,breaklinks=true,pdftex]{hyperref}
\usepackage{bm}
\usepackage{color}
\usepackage{amsmath, amsthm,amssymb}
\usepackage{multirow}
\usepackage{float}
\usepackage{wrapfig,lipsum,booktabs}
\modulolinenumbers[1]

\journal{CAGD 2023}

\biboptions{authoryear}

\begin{document}

\begin{frontmatter}

\title{D-Net: Learning for Distinctive Point Clouds by Self-Attentive Point Searching and Learnable Feature Fusion}


\author[first]{Xinhai Liu}
\ead{adlerxhliu@tencent.com}
\author[second]{Zhizhong Han}
\ead{h312h@wayne.edu}
\author[first]{Sanghuk Lee}
\ead{sanghuklee16@gmail.com}
\author[last]{Yan-Pei Cao}
\ead{caoyanpei@gmail.com}
\author[first]{Yu-Shen Liu\corref{cor}}
\ead{liuyushen@tsinghua.edu.cn}
\cortext[cor]{Corresponding author}
\address[first]{School of Software, Tsinghua University, Beijing, China}
\address[second]{Department of Computer Science, Wayne State University, USA}
\address[last]{Y-tech, Kuaishou Technology, China}

\begin{abstract}
    Learning and selecting important points on a point cloud is crucial for point cloud understanding in various applications.
    Most of early methods selected the important points on 3D shapes by analyzing the intrinsic geometric properties of every single shape, which fails to capture the importance of points that distinguishes a shape from objects of other classes, i.e., the distinction of points.
    To address this problem, we propose D-Net (Distinctive Network) to learn for distinctive point clouds based on a self-attentive point searching and a learnable feature fusion.
    Specifically, in the self-attentive point searching, we first learn the distinction score for each point to reveal the distinction distribution of the point cloud.
    After ranking the learned distinction scores, we group a point cloud into a high distinctive point set and a low distinctive one to enrich the fine-grained point cloud structure.
    To generate a compact feature representation for each distinctive point set, a stacked self-gated convolution is proposed to extract the distinctive features.
    Finally, we further introduce a learnable feature fusion mechanism to aggregate multiple distinctive features into a global point cloud representation in a channel-wise aggregation manner.
    The results also show that the learned distinction distribution of a point cloud is highly consistent with objects of the same class and different from objects of other classes.
    Extensive experiments on public datasets, including ModelNet and ShapeNet part dataset, demonstrate the ability to learn for distinctive point clouds, which helps to achieve the state-of-the-art performance in some shape understanding applications. 
\end{abstract}


\end{frontmatter}


\section{Introduction}
Point cloud captured by 3D scanners has attracted extensive research interest in various point cloud understanding applications, such as 3D object classification \cite{qi2017pointnet,qi2017pointnet++,wang2019dynamic,liu2019relation}, detection \cite{shi2019pointrcnn,chen2019clusternet,lang2019pointpillars} and segmentation \cite{markhamrandla,yang2019learning,wang2019exploiting}.
Unlike regular 2D images and 3D voxels with fixed local spatial distribution, the point cloud is irregular and unordered, consisting of 3D coordinates and some additional attributes, e.g., color, normal, reflectance, etc.
To avoid processing unstructured point sets, it is intuitive to first convert point clouds into regular views \cite{chen2017multi,roveri2018network} or voxels \cite{zhou2018voxelnet,meng2019vv} and then apply traditional convolutional neural networks to complete the recognition process.
However, due to lacking depth information, 2D view-based methods are limited in multiple applications like semantic segmentation, and 3D voxel-based methods require high memory and computational cost.
\begin{figure}[tp]
    \centering
    \includegraphics[width=15cm]{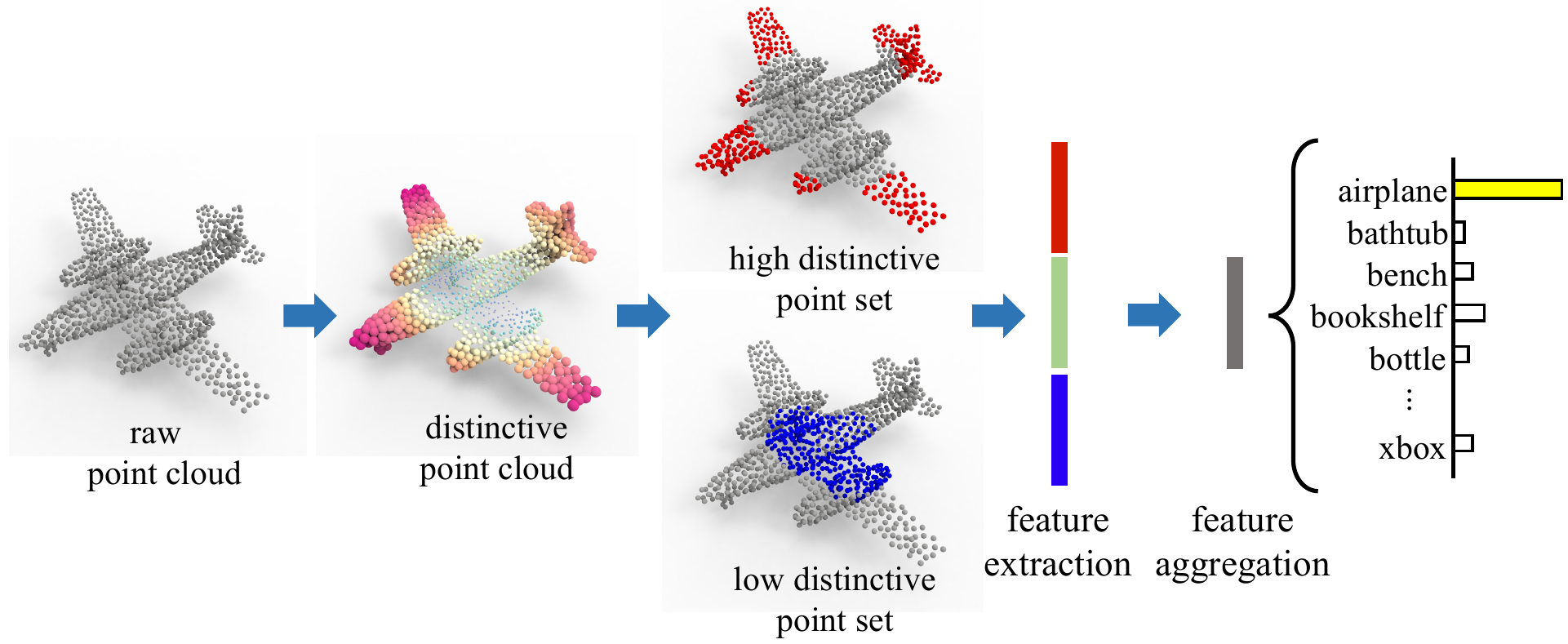}
    \caption{The learning for distinctive point clouds.
    Specifically, for a raw point cloud, we first build the distinctive point cloud by learning a distinction score for each point, where the red points with a large radius indicate a high distinction and the blue points with a small radius indicate a low distinction.
    By ranking the distinction scores, we respectively select the high distinctive point set and low distinctive point set to enrich fine-grained details.
    Then, we extract the corresponding feature representation of each distinctive point set, as well as the raw point cloud.
    The extracted features of three point sets are aggregated into a global point cloud representation, which distinguishes a 3D shape from others of different classes.}
    \label{fig:idea}
\end{figure}

Many subsequent improvements \cite{li2018pointcnn,wang2019dynamic,liu2020lrc,liu2019relation} mainly focus on capturing local or global structures within point clouds and have achieved better performances.
They usually treat all points of a point cloud equally.
However, the important regional points on point clouds and their correlation are not well measured and detected for learning the global shape representation for distinguishing shapes from others in recognition tasks.
In this paper, we found that learning and selecting distinctive points from point clouds are important to distinguish each shape from shapes of other classes.
We follow the definition of the distinction of points on a 3D object, as proposed by Shilane and Funkhouser \cite{shilane2007distinctive}, for learning and selecting the important points, where the distinction of every point in an object is defined as how useful it is for distinguishing the object from others of different classes.
Most prior methods \cite{lee2005mesh,shilane2006selecting} have selected the important points on shapes by analyzing the intrinsic geometric properties of every single shape, which lacks to capture the importance of points that distinguish each shape from objects of other types. 
This common strategy is first to extract the hand-crafted shape features from the local neighborhood and then obtain the distinctiveness of regional points based on the difference between their features to others.
To alleviate the limitation of hand-crafted shape features, a few approaches \cite{shu2018detecting,nezhadarya2019adaptive,zheng2019pointcloud} have been proposed to learn to detect the points of interest or salient points, which often consider how unique and visible a point is relative to other points within the same object. 

In contrast, the distinction of a point considers how common and unique the point is relative to objects of other types in a dataset. 
In addition, the above-mentioned methods also lack to explore the spatial correlation among multiple distinctive point sets of a point cloud.
To address this problem, we propose D-Net (Distinctive Network) to explore both the local structures within points and the spatial correlation among multiple distinctive point sets of a point cloud.
As shown in Fig. \ref{fig:idea}, we first learn the distinction score for each point to reveal the point distinction distribution. 
By ranking the learned distinction scores, we group each point cloud into a high distinctive point sets and a low distinctive point sets. 
Then, to extract the feature representation of each distinctive point set, a stacked self-gated convolution is proposed for extracting the distinctive features. 
Finally, a feature aggregation operation is designed for fusing multiple distinctive point set features.
In feature fusion, the pooling operation has been widely adopted by existing works to combine multiple features with deep neural networks, but simply pooling disregards a lot of content information and the spatial correlation among different distinctive point set features, which limits the discriminability of learned global shape representation.
To address this issue, we further introduce a learnable feature fusion mechanism to aggregate the distinctive point set features into a global point cloud representation in a channel-wise aggregation manner.
Our main contributions are summarized as follows.
\begin{itemize}
    \item We design a self-attentive point searching module to capturing a distinction score for each point, which provides a primary basis of the ranking of points though the back-propagation of gradients.
    \item To effectively capture the features of point sets, we propose a self-gated convolution module, which enables the information forgetting in different abstraction layers by introducing the gate structure.
    \item To aggregate the multiple features of distinctive point sets in a channel-wise way, we introduce a learnable gate fusion strategy to adaptively aggregate multiple distinctive features.

\end{itemize}

\section{Related work}
\noindent\textbf{Learning on Point Clouds.}
Recently, deep networks have achieved promising performances in point cloud analysis. 
However, due to the irregular and unordered data representation, there are still some challenges learning from raw point clouds.
PointNet \cite{qi2017pointnet} is the prior work that applies deep model directly on point clouds.
Specifically, a point-wise feature is first extracted for each point independently, and all features are then aggregated into the global representation with a max-pooling operation.
However, PointNet missed the important local region structures of point clouds.
PointNet++ \cite{qi2017pointnet++} employed a hierarchical strategy to capture the local structures inside small point clusters.
To mitigate the impact of point cloud irregularities, some studies utilize scalable indexing techniques to build regular structures on points.
OctNet \cite{riegler2017octnet} hierarchically divided points into regular grids with unbalanced octrees, and extracted the corresponding feature for each leaf node.
Kd-Net \cite{klokov2017escape} performed multiplicative transformations according to the subdivisions of point clouds based on kd-trees. 

Influenced by traditional convolutional neural networks, some other methods focus on building graphs in local regions and applying CNN-like operations to capture local structures.
PointCNN \cite{li2018pointcnn} used the canonical order of local region points and applied a typical convolution operator on transformed features.
SPGraph \cite{landrieu2018large} partitioned large scale points into geometrically homogeneous elements with building a super-point graph.
RS-CNN \cite{liu2019relation} also extended typical CNN to irregular configuration, which encoded the geometric relation of points to achieve contextual shape-aware learning of point cloud.
DGCNN \cite{wang2019dynamic} applied an EdgeConv operation to capture the local region structures with dynamic graph updating.
These approaches mainly focus on capturing local or global structures of point clouds, which ignores the difference of point importance in distinguishing shape classes.
To address this problem, we propose a self-attentive point searching module to capture the point importance and explore fine-grained structures within multiple distinctive point sets.

\noindent\textbf{Distinction Detection on Point Clouds.}
The distinction, or distinctiveness, was first introduced by Shilane and Funkhouser \cite{shilane2006selecting,shilane2007distinctive}.
These methods rely on extracting local shape descriptors and obtaining the distinctiveness of each local region by comparing the difference between pairs of shape descriptors. One hand-crafted descriptor is usually designed for a specific task, which cannot generalize well to another tasks. And it is difficult to find the best combination of existing hand-crafted descriptors for the current task.
To alleviate the limitation of hand-crafted shape descriptors, several methods \cite{shu2018detecting,li2020unsupervised,song2018distinction} employed learning-based strategies to capture shape distinction by back-propagation optimization.
Besides 3D shapes, several approaches have been developed to extract the discriminative regions from images.
Similar to distinction, saliency has also been explored by \cite{gal2006salient,wang2018tracking}, which considered how unique and visible of regions within the same object.
More recently, CP-Net \cite{nezhadarya2019adaptive} performed adaptive down-sampling with considering point importance by making channel-wise statistics on point features.
In \cite{zheng2019pointcloud}, a way of characterizing critical points and segments to build point-cloud saliency maps was proposed.
However, existing approaches only focus on extracting high distinctive points, which ignores the spatial correlation information among different distinctive regions.
The distinctive regions and their spatial correlations usually provide high-level information for distinguishing every object from others of different classes \cite{shilane2007distinctive}.

Inspired by the region distinction \cite{shilane2007distinctive}, we propose an end-to-end network, called D-Net, to learn point distinction directly from raw point clouds.
However, different from the original distinction region, we conduct the distinction detection in a learning-based manner.
So, the learned distinctive or important points are relative to the training task, such as shape classification and shape part segmentation.
In D-Net, we pay attention to different distinctive regional point sets that contain fine-grained detail information of point clouds and distinguish a shape from objects of a different class.
We rely on self-attentive point searching to learn different distinctive point sets and capture the corresponding feature representations with stacked self-gated convolution.
Moreover, the final global representation of the point cloud is aggregated with a learnable feature fusion mechanism by leveraging the correlation of distinctive point sets.

\begin{figure}[t]
    \centering
    \includegraphics[width=16cm]{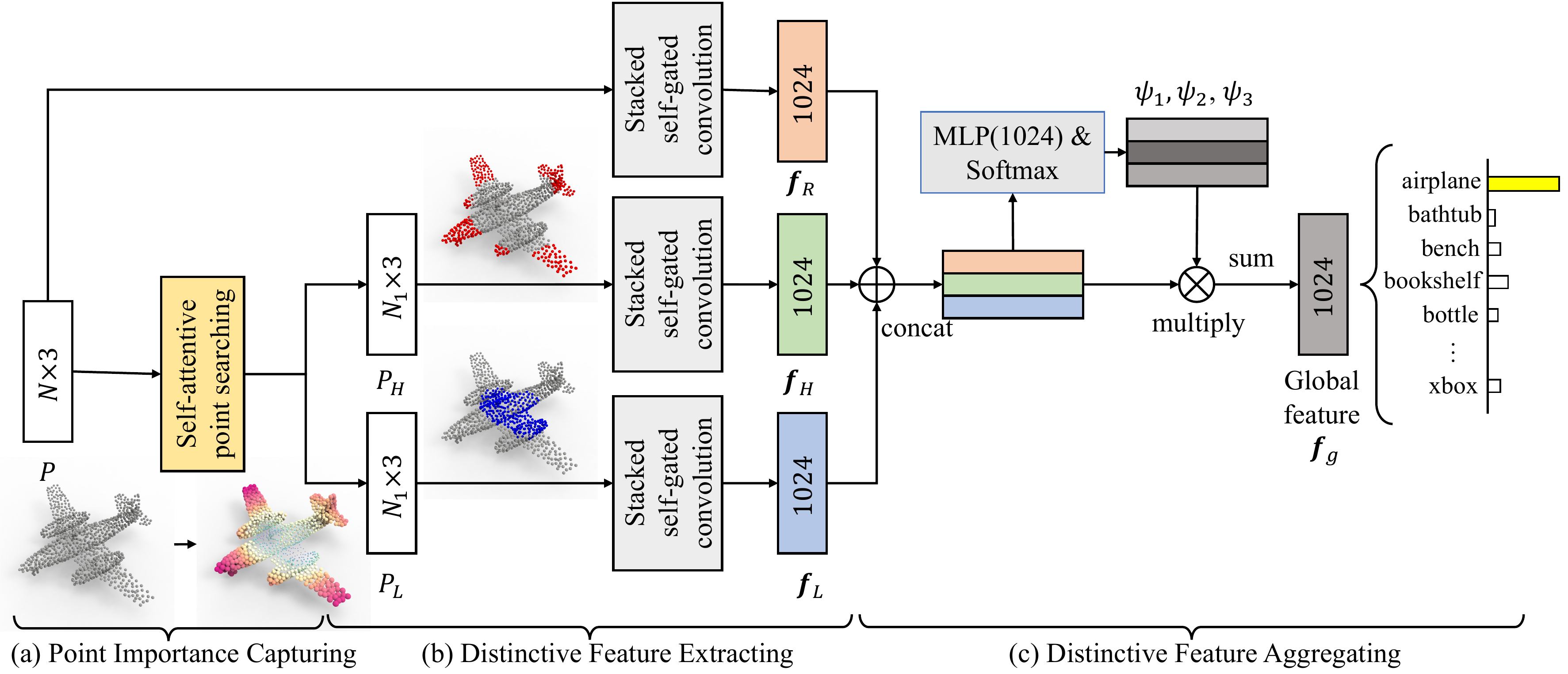}
    \caption{Our D-Net architecture. Our D-Net has three components, including the point importance capturing, the distinctive feature extracting, and the distinctive feature aggregating. 
    Specifically, the point importance capturing relies on self-attentive point searching to learn a distinction score for each point. 
    By ranking the distinction scores of points, we select out the high distinctive point set $P_H$ and the low distinctive point set $P_L$ to enrich the fine-grained structures of point clouds.
    With stacked self-gated convolution, we extract the feature representation for each distinctive point set.
    We resort to a learnable gate for feature aggregation to effectively aggregate point set features $\bm{f}_R$, $\bm{f}_H$, and $\bm{f}_L$ into a compact global point cloud representation $\bm{f}_g$.
    The final global representation can be applied to point cloud analysis tasks, including shape classification and shape part segmentation.}
    \label{fig:main_frame}
\end{figure}

\section{The D-Net method}
The architecture of our D-Net is illustrated in Figure \ref{fig:main_frame}, which consists of three components: the point importance capturing, the distinctive feature extracting, and distinctive feature aggregating. 
We will illustrate the network details of our D-Net as follows.

\noindent\textbf{Point Importance Capturing.}
Each point in a point cloud has different importance.
To capture structures within the irregular point cloud, we should consider not only the raw point cloud itself but also some fine-grained details within distinctive point sets.
We propose a novel self-attentive point searching layer to enrich point cloud information by taking point distinction into consideration.
The high distinctive point set and the low distinctive point set are selected out to enhance the point cloud representation learning.
The enriched details from distinctive point sets can help to boost point cloud analysis tasks, such as shape classification performances.

\noindent\textbf{Distinctive Feature Extracting.}
With multiple distinctive point sets, we resort to a CNN-like operation with gated aggregation to capture the local region contexts within each distinctive point set.
To further exploit the local structures of the point set, the SGC (Self-Gated Convolution) is employed in the point set representation learning, which relies on building graphs that dynamically compose and update the feature representation of each point within a local region. 
The feature representation of each distinctive point set is extracted by aggregating the different point features with stacked self-gated convolution.

\noindent\textbf{Distinctive Feature Aggregating.}
Based on the obtained feature representations of distinctive point sets, we introduce a learnable feature fusion mechanism to integrate these features into a compact global point cloud representation.
We calculate channel-wise gate weights for each corresponding feature to utilize the spatial correlation among different distinctive features.
Afterward, the multiple distinctive set features are aggregated into the final global point cloud representation.
\begin{figure}[tp]
    \centering
    \includegraphics[width=16cm]{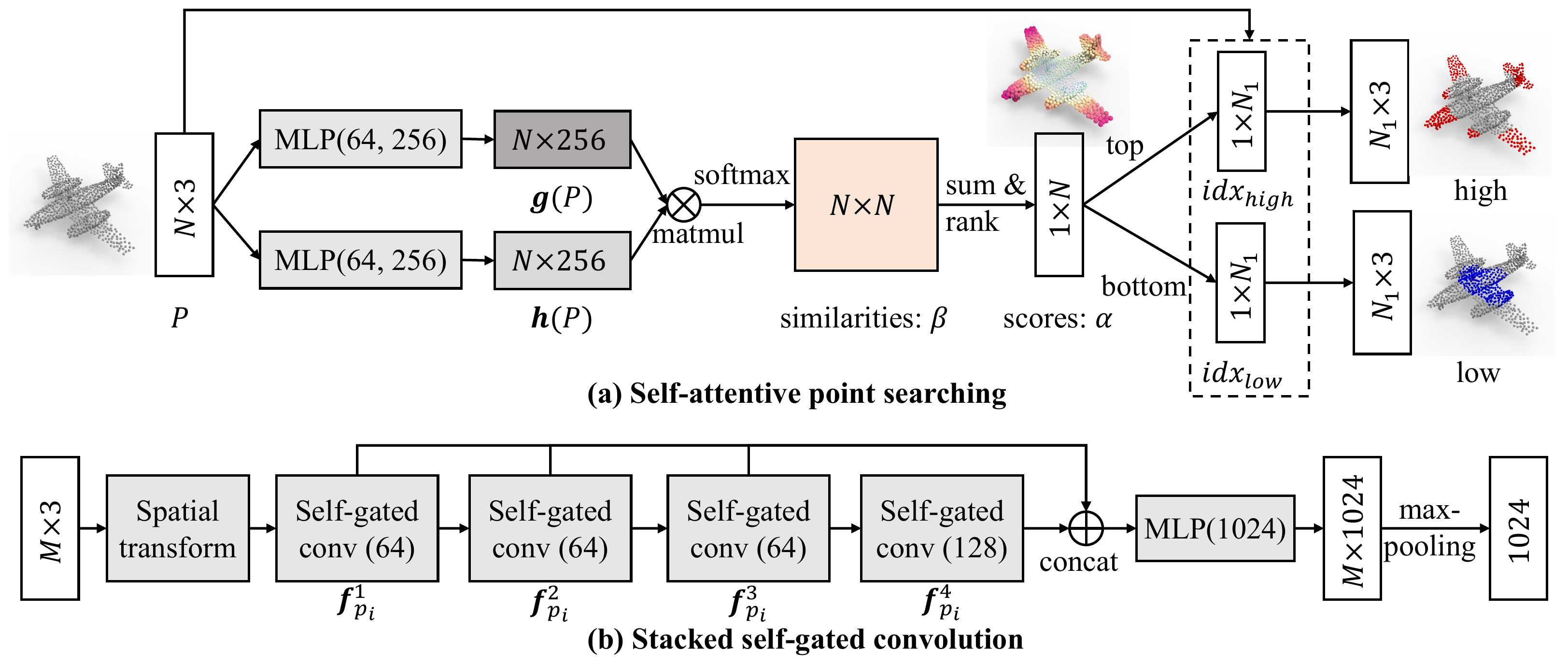}
    \caption{The proposed modules, including (a) self-attentive point searching module and (b) stacked self-gated convolution module.}
    \label{fig:p_s}
\end{figure}

\subsection{Self-attentive Point Searching}
The existing point cloud based learning approaches usually treat all points of a point cloud equally, without distinguishing the point distinction that plays an essential role in distinguishing a shape from objects of a different class \cite{shilane2006selecting,shilane2007distinctive,li2020unsupervised}.
As shown in Fig. \ref{fig:idea}, the point sets in the point cloud can be roughly divided into the high distinctive and the low distinctive point sets.
In this paper, besides the raw point cloud, we incorporate the high distinctive point set and the low distinctive point set to enrich the discriminative context of point clouds.
With such incorporated contextual information, D-Net can capture the detailed point cloud structure information.

As illustrated in Fig. \ref{fig:p_s} (a), given a point cloud $P=\{p_{i=1}^N\}$ with $N$ points, where $p_i \in \mathbb{R}^3$ denotes the 3D coordinates $(x_i,y_i,z_i)$ of the $i$-th point.
To obtain the distinction score of points, we first transform the point cloud $P$ into two feature space $\bm{g}, \bm{h} \in \mathbb{R}^D$ to calculate the bilinear similarity between points, where $\bm{g}(P) = \bm{W_g}P$, $\bm{h}(P) = \bm{W_h}P$,

\begin{equation}
\begin{split}
    \beta_{j,i} = \frac{exp(s_{ij})}{\sum_{i=1}^{D}{s_{ij}}}, \; &\text{where} \; s_{ij} = \bm{g}(p_i)^T\bm{h}(p_j),\; \alpha_i =  \sum_{j=1}^{N}{\beta_{j,i}}, \\
    &idx_{high} = \mathop{top}_{N_1}(sort(\{\alpha_{i}\})), \\ 
    &idx_{low} = \mathop{bottom}_{N_1}(sort(\{\alpha_{i}\}))
\end{split}
\end{equation}

Specifically, $\bm{W_g}$ and $\bm{W_h}$ are learnable parameters of Multi-Layer Perceptrons (MLPs). $\beta_{j,i}$ evaluates the bilinear similarity between the $i$-th point $p_i$ and the $j$-th point $p_j$ in the feature space and $\alpha_{i}$ denotes the self-attentive score of the $i$-th point $p_i$.
According to the ranking of self-attentive scores, we search out the index $idx_{high}$ of top $N_1$ points as the high distinctive point set and $idx_{low}$ of bottom $N_1$ points as the low distinctive point set, respectively.
With the point index $idx_{high}$ and $idx_{low}$, we pick out the corresponding points to establish the high distinctive point set $P_H$ and low distinctive point set $P_L$.
As such, the contextual information of the raw point cloud is enriched with two selected distinctive point sets.
Note that selecting the low distinctive point set $P_L$ in this step aims to make up for the neglecting structural information of point clouds.

\subsection{Stacked Self-gated Convolution}
Based on the enriched distinctive point sets, we rely on a distinctive feature extracting (DFE) module with the self-gated CNN to extract the feature representation for each distinctive point set.
The architecture of DFE module is illustrated in Fig. \ref{fig:p_s} (b).
Following previous approaches \cite{qi2017pointnet,wang2019dynamic}, we first employ a transform layer to align an input point set to a canonical space by applying an estimated $3 \times 3$ matrix.
The $k$ neighboring points are used in the estimation of $3 \times 3$ matrix.
We will extract the local structures based on the transformed point set with CNN-like operations.

\noindent\textbf{Self-gated Convolution.}
Local structure information has been proven to be important in learning the point cloud representation.
In order to capture local structures, we propose one novel self-gated CNN to capture the local region context around each point.
Similar to \cite{qi2017pointnet++,wang2019dynamic}, we first use $k$-nearest neighbor ($k$-NN) to build a local region $P_r \subset \mathbb{R}^3$, with $k$ surrounding points as its neighbors $p_j \in \mathcal{N}(p_i)$ of $p_i$.
Our goal is to learn an inductive representation $\bm{f}_{P_r}$ of this local region, which should discriminatively encode the underlying shape information.

\begin{figure}
    \centering
    \includegraphics[width=10cm]{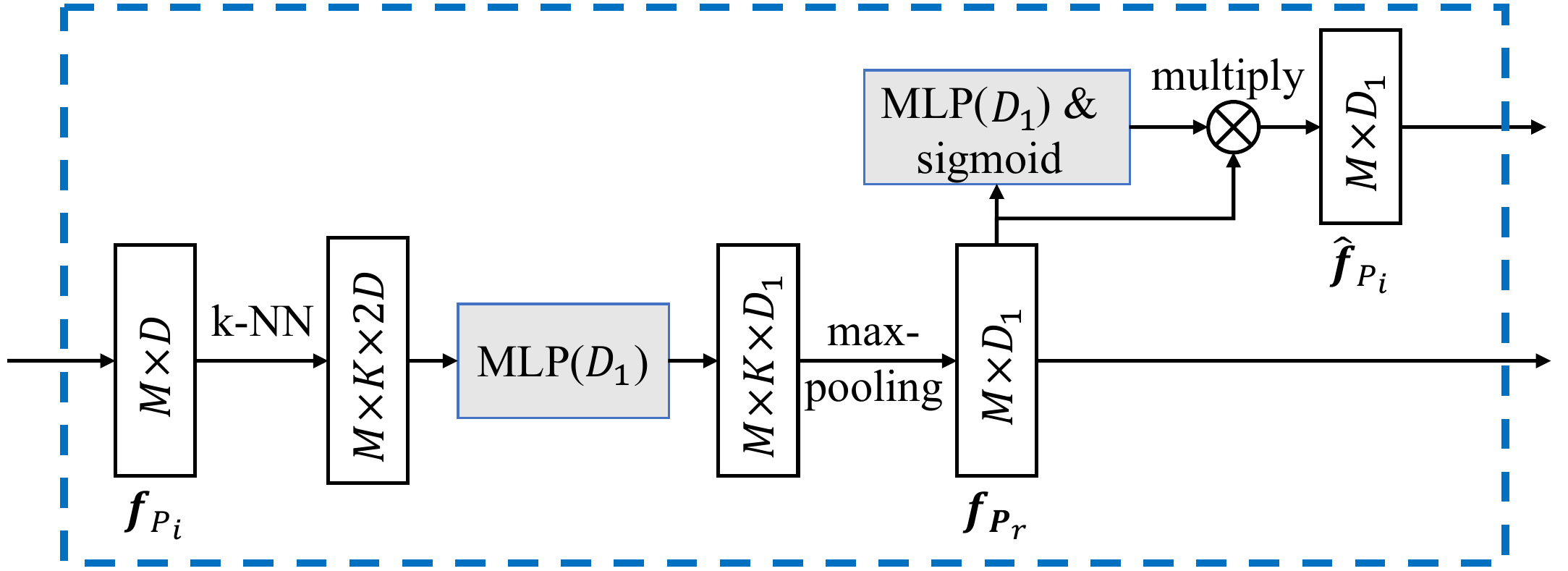}
    \caption{Self-gated convolution.}
    \label{fig:conv}
\end{figure}

To this end, we formulate a general convolutional operation as
\begin{equation}
\begin{aligned}
    &\bm{f}_{P_r} = \sigma(\mathcal{A}(\{ \mathcal{T}(\bm{h}_{ij}), \forall p_j \})), \forall p_j \in \mathcal{N}(p_i), \\
    &\bm{h}_{ij} = \bm{f}_{p_i} \oplus (\bm{f}_{p_j} - \bm{f}_{p_i}),
\end{aligned}
\end{equation}
where $\bm{f}$ is a feature vector, $\oplus$ is the concatenation operation and $\bm{h}_{ij}$ is the relation description of feature vectors between two points. 
Here $\bm{f}_{P_r}$ is obtained by first transforming the features of all the points in $\mathcal{N}(p_i)$ with function $\mathcal{T}$ , and then aggregating them with function $\mathcal{A}$ followed by a nonlinear activator $\sigma$. 
In this formulation, the two functions $\mathcal{A}$ and $\mathcal{T}$ are the key to $\bm{f}_{P_r}$. 
That is, the permutation invariance of point set can be achieved only when $\mathcal{A}$ is symmetric (e.g., max-pooling) and $\mathcal{T}$ is shared over each point in $\mathcal{N}(p_i)$.
As illustrated in Fig \ref{fig:conv}, $\mathcal{T}$ is a MLP function, $\mathcal{A}$ is a max-pooling operation and $\sigma$ is a ReLU layer.
Through this convolution operation, we update the point feature $\bm{f}_{P_i}$ with the regional feature $\bm{f}_{P_r}$, which contains structural information of local regions.

By stacking $T$ self-gated convolution modules, we can obtain multiple point features $\{\bm{f}^1_{p_i},\cdots, \bm{f}^t_{p_i}, \cdots, \bm{f}^T_{p_i} \}$ of $p_i$ in different depth of network as shown in Fig \ref{fig:p_s} (b).
In particular, we also dynamically update the graph of local regions as in \cite{wang2019dynamic}.
To obtain the feature representation of each distinctive point set, we first integrate these features into one consistent point representation.
A novel self-gate is proposed to integrate these multiple point features.
We formulate the self-gated aggregation process as 
\begin{equation}
\begin{split}
    &\theta^{t}_{P_i}  = sigmoid(MLP(\bm{f}^t_{p_i})), \\
    &\bm{\hat{f}}^t_{p_i}  = \theta^{t}_{P_i} \otimes  \bm{f}^t_{p_i}, \\
    &\bm{\hat{f}}_{p_i} = \bm{\hat{f}}^1_{p_i} \oplus \cdots \oplus \bm{\hat{f}}^t_{p_i} \oplus \cdots \oplus \bm{\hat{f}}^T_{p_i},\\
    &\bm{f} = max(\{\bm{\hat{f}}_{p_i}\}),
\end{split}
\end{equation} 
where $\theta^{t}_{p_i}$ is the product factor of the self-gate and $\bm{f}_{p_i}^t$ is the gated point feature at $t$-th depth of the network.
$\otimes$ denotes multiply operation and $\oplus$ denotes the concatenation operation.
$\bm{\hat{f}}_{p_i}$ is the aggregated point feature from multiple depth features.
With a max-pooling, the corresponding feature of each distinctive point set is obtained, including the point cloud feature $\bm{f}_{P}$, the high distinctive point set feature $\bm{f}_{P_H}$ and the low distinctive point set feature $\bm{f}_{P_L}$.

\subsection{Learnable Feature Fusion}
Our goal is to aggregate the obtained point cloud feature $\bm{f}_{P}$, the high distinctive point set feature $\bm{f}_{P_H}$ and the low distinctive point set feature $\bm{f}_{P_L}$ into a global point cloud representation.
To fuse information of different distinctive point sets, we propose a non-linear and data-adaptive mechanism to explore the correlation among distinctive point sets.
By applying a \textit{softmax} on newly mapped descriptors, we formulate the fusion process as 
\begin{equation}
\begin{aligned}
    \omega_1 = MLP(\bm{f}_{P}), \omega_2 = MLP(\bm{f}_{P_H}), \omega_3 = MLP(\bm{f}_{P_L}),\\
    \psi_{1,c} = \frac{exp(\omega_{1,c})}{exp(\omega_{1,c})+exp(\omega_{2,c})+exp(\omega_{3,c})},\\
    \psi_{2,c} = \frac{exp(\omega_{2,c})}{exp(\omega_{1,c})+exp(\omega_{2,c})+exp(\omega_{3,c})},\\
    \psi_{3,c} = \frac{exp(\omega_{3,c})}{exp(\omega_{1,c})+exp(\omega_{2,c})+exp(\omega_{3,c})}, 
\end{aligned}
\end{equation}
where $\omega_1, \omega_2, \omega_3 \in \mathbb{R}^{1024}$ are new mapped descriptors that are transformed from channel-wise descriptors $\bm{f}_{P}$, $\bm{f}_{P_H}$ and $\bm{f}_{P_L}$ with different MLPs.
$\psi_{1,c}$, $\psi_{2,c}$ and $\psi_{3,c}$ are the fusion weights of point set features, where $\psi_{1,c} + \psi_{2,c} + \psi_{3,c} = 1$. In particular, the symbol $c$ indicates that the learnable feature fusion is a channel-wise operation, which enables the exploration of general patterns between channels.

The ultimate global point cloud feature $\bm{f}_{g}$ can be fused by a weighted sum operation as
\begin{equation}
    \bm{f}_{g:c} = \psi_{1,c} \cdot \bm{f}_{P:1,c} + \psi_{2,c} \cdot \bm{f}_{P_H:2,c} + \psi_{1,c} \cdot \bm{f}_{P_L:1,c},
\end{equation}
where $\bm{f}_{g:c}$ denotes the $c$-th channel value of $\bm{f}_{g}$. The learned global shape representation $\bm{f}_{g}$ can be applied to various point cloud analysis applications, such as shape classification and shape part segmentation.

\subsection{Training Loss}
In the representation learning of point clouds, we focus on capturing the discriminative point cloud features, which can be applied to multiple point analysis applications, including point cloud classification and shape part segmentation.
Following the strategy of previous point cloud recognition methods, we train our approach in an end-to-end manner with the cross-entropy loss.
For example, in the point cloud classification task, we classify the extract global feature $\bm{f}_g$ into one of $C$ shape classes by a softmax function layer.
The softmax function outputs the classification probabilities $\bm{p}$, such that each value $\{\bm{p}(c), c \in [1, C] \}$ indicates the probability under the current class.
The objective function of $\mathcal{L}_{ce}$ is the cross entropy between $\bm{p}$ and the ground-truth probability $\bm{p}'$ ,
\begin{equation}
    \mathcal{L}_{ce}(\bm{p}, \bm{p}') = -\sum_{c \in [1, C]}{\bm{p}'(c)log \bm{p}(c)}.
\end{equation}

Similar to point cloud classification, the shape part segmentation can be regarded as a low-level point-wise classification task.
Therefore, we also employ the cross entropy loss in the shape part segmentation task.
\section{Experiments}
In this section, we arrange comprehensive experiments to validate the proposed D-Net.
First, we investigate how some key parameters affect the performance of D-Net.
And then, we evaluate D-Net in shape classification and shape part segmentation, respectively.
We also conducted an ablation study to evaluate the effectiveness of each proposed module under the shape classification benchmark.
In addition, we visualize some experimental results to show the performance of our approach intuitively.
Finally, we do a model complexity analysis to show the effectiveness of our D-Net.

\begin{table}
    \centering
    \scalebox{0.9}{
    \begin{tabular}{lccccc}
        \hline 
        Methods                                   &Supervised  &Input           &\#Points       &Acc. \\ \hline
        PointNet \cite{qi2017pointnet}             &yes   &xyz                       &1k             &89.2 \\
        PointNet++ \cite{qi2017pointnet++}          &yes  &xyz                       &1k             &90.7 \\
        Kd-Net(depth=10) \cite{klokov2017escape}    &yes &xyz                       &1k             &90.6 \\
        KC-Net \cite{shen2018mining}              &yes  &xyz                        &1k             &91.0 \\
        ShapeContextNet \cite{xie2018attentional}   &yes  &xyz                       &1k             &90.0 \\
        LRC-Net \cite{liu2020lrc}                    &yes  &xyz                        &1k             &93.1 \\
        PointCNN \cite{li2018pointcnn}             &yes   &xyz                         &1k             &91.7 \\
        PCNN \cite{2018atzmonpcnn}                 &yes   &xyz                          &1k             &92.3 \\    
        DGCNN \cite{wang2019dynamic}                 &yes  &xyz                        &1k             &92.9 \\
        Point2Sequence \cite{liu2019point2sequence} &yes &xyz                           &1k             &92.6 \\
        A-CNN \cite{komarichev2019cnn}              &yes &xyz                        &1k             &92.6 \\
        RS-CNN \cite{liu2019relation}               &yes &xyz                            &1k             &92.9 \\
        SO-Net \cite{li2018so}                      &yes &xyz                          &2k             &90.9 \\
        Kd-Net(depth=15) \cite{klokov2017escape}    &yes &xyz                        &32k            &91.8 \\
        CP-Net \cite{nezhadarya2019adaptive}        &yes &xyz                        &1k             &92.4 \\ 
        Grid-GCN \cite{xu2020grid}                  &yes &xyz                       &16k             &93.1 \\
        PACov \cite{xu2021paconv}                   &yes &xyz                       &1k              &93.6 \\
        PointASNL \cite{yan2020pointasnl}           &yes &xyz                       &1k              &93.2 \\
        PointTransformer \cite{zhao2021point}       &yes &xyz                       &1k              &\textbf{93.7} \\
        O-CNN \cite{wang2017cnn}                    &yes &xyz+nor.                   & -             &90.6 \\
        Spec-GCN \cite{wang2018local}               &yes &xyz+nor.                  &10k            &91.8 \\
        PointNet++ \cite{qi2017pointnet++}          &yes &xyz+nor.                 &5k             &91.9 \\
        SpiderCNN \cite{xu2018spidercnn}            &yes &xyz+nor.                  &5k             &92.4 \\
        \textbf{D-Net(Ours)}                          &yes      &xyz            &1k    &93.2 \\      
        \textbf{D-Net(Ours)}                           &yes    &xyz+nor.             &1k    &93.3 \\ \hline                     
        LGAN     \cite{achlioptas2017learning}     &no  &xyz                            &2k               &85.7  \\
        FoldingNet  \cite{yang2018foldingnet}      &no  &xyz                         &2k              &88.4  \\
        MAP-VAE     \cite{han2019multi}            &no &xyz                         &2k               &90.2 \\
        UDDR \cite{li2020unsupervised}             &no &xyz                         &1k    &88.1\\
        L2G-AE   \cite{liu2019l2g}                 &no  &xyz                          &1k              &90.6 \\ 
        \textbf{D-Net++(Ours)}                      &no  &xyz      &1k     &\textbf{90.9}  \\ \hline
    \end{tabular}}
    \caption{Shape classification results (\%) under ModelNet40.}
    \label{table:cls}
\end{table}

 \begin{figure}[t]
    \centering
    \includegraphics[width=\textwidth]{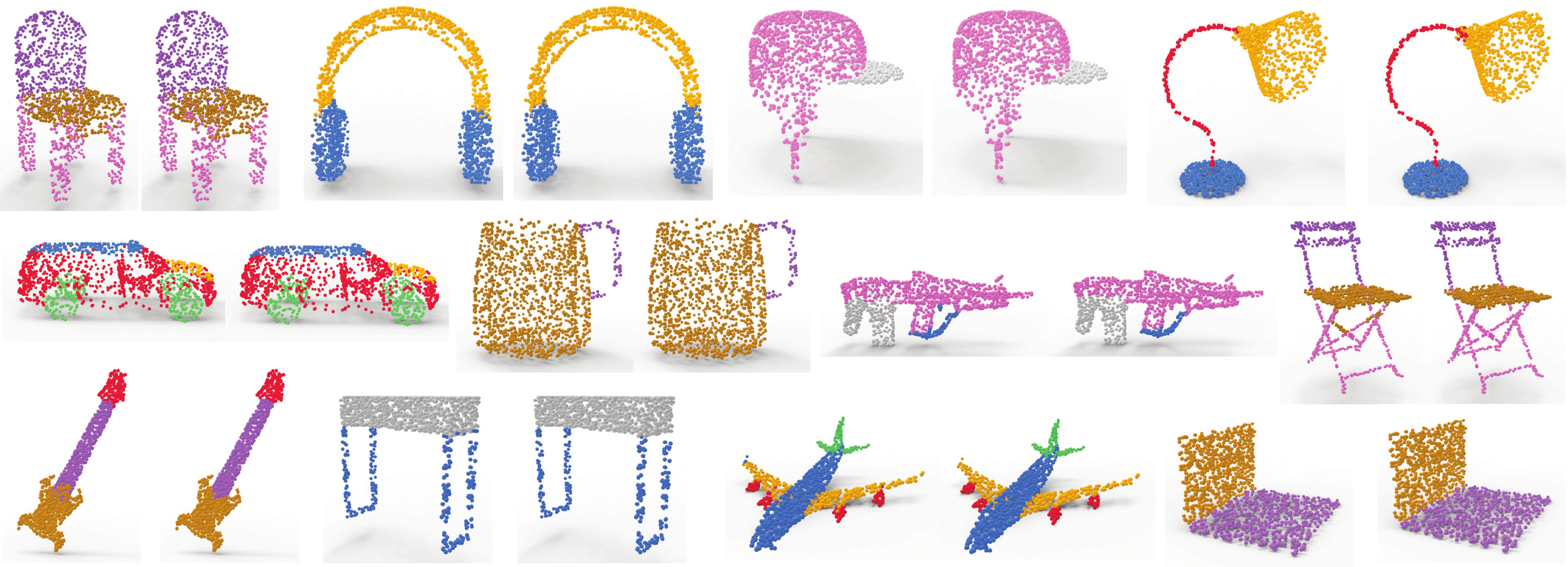}
    \caption{The visualization of shape part segmentation. For each pair of 3D shapes, the left one indicates the ground-truth shape segmentation, and the right one is the predicted shape segmentation. 
    In each shape, each color denotes a specific semantic part label.}
    \label{fig:sem}
\end{figure}
\begin{table}[t]
    \resizebox{\textwidth}{!}{
    \begin{tabular}{l|c|ccccccccccccccccccc}
    \hline
    \multirow{2}{*}{Methods}&
        \multicolumn{1}{c|}{\multirow{2}{*}{Mean IoU}} &
        \multicolumn{16}{c}{Intersection over Union (IoU)}\\
        & \multicolumn{1}{c|}{} 
        & \multicolumn{1}{c}{air.}
        & \multicolumn{1}{c}{bag}
        & \multicolumn{1}{c}{cap}
        & \multicolumn{1}{c}{car}
        & \multicolumn{1}{c}{cha.}
        & \multicolumn{1}{c}{ear.}
        & \multicolumn{1}{c}{gui.}
        & \multicolumn{1}{c}{kni.}
        & \multicolumn{1}{c}{lam.}
        & \multicolumn{1}{c}{lap.}
        & \multicolumn{1}{c}{mot.}
        & \multicolumn{1}{c}{mug}
        & \multicolumn{1}{c}{pis.}
        & \multicolumn{1}{c}{roc.}
        & \multicolumn{1}{c}{ska.}
        & \multicolumn{1}{c}{tab.}
        \\ \hline
    \# SHAPES  & &2690 &76 &55 &898 &3758 &69 &787 &392 &1547 &451 &202 &184 &283 &66 &152 &5271 \\
    PointNet \cite{qi2017pointnet}			           &83.7           &83.4           &78.7           &82.5           &74.9           &89.6           &73.0&91.5&85.9&80.8&95.3&65.2&93.0&81.2&57.9&72.8&80.6 \\
    Kd-Net	\cite{klokov2017escape}			            &82.3           &80.1           &74.6           &74.3           &70.3           &88.6           &73.5&90.2&87.2&81.0&94.9&57.4&86.7&78.1&51.8&69.9&80.3 \\
    ShapeContextNet \cite{xie2018attentional}           &84.6           &83.8           &80.8           &83.5           &79.3           &90.5           &69.8 &91.7 &86.5 &82.9 &96.0&69.2 &93.8 &82.5&62.9&74.4 &80.8 \\
    KCNet   	\cite{shen2018mining}		          &84.7           &82.8           &81.5           &86.4           &77.6           &90.3           &76.8&91.0&87.2&84.5&95.5&69.2&94.4&81.6&60.1&75.2&81.3 \\
    DGCNN \cite{wang2019dynamic}                       &85.1           &84.2           &83.7           &84.4           &77.1           &90.9           &78.5&91.5 &87.3 &82.9 &96.0 &67.8 &93.3 &82.6 &59.7 &75.5 &82.0	\\
    SO-Net 	\cite{li2018so}			               &84.9           &82.8           &77.8           &88.0           &77.3           &90.6           &73.5&90.7&83.9&82.8&94.8&69.1&94.2&80.9&53.1&72.9&83.0 \\
    A-CNN \cite{komarichev2019cnn}                     &86.1           &84.2           &84.0           &88.0           &79.6           &\textbf{91.3}  &75.2&91.6&87.1&85.5&95.4&75.3&94.9&82.5&\textbf{67.8}&77.5&83.3 \\
    RS-CNN \cite{liu2019relation}                     &\textbf{86.2}  &83.5           &84.8           &88.8           &79.6           &91.2           &81.1 &91.6 &\textbf{88.4} &\textbf{86.0} &96.0 &73.7 &94.1 &83.4 &60.5 &77.7 &\textbf{83.6} \\
    PointNet++  \cite{qi2017pointnet++}		           &85.1           &82.4           &79.0           &87.7           &77.3           &90.8           &71.8&91.0&85.9&83.7&95.3&71.6&94.1&81.3&58.7&76.4&82.6 \\
    PointCNN \cite{li2018pointcnn}                    &86.1           &84.1           &\textbf{86.5}  &86.0           &\textbf{80.8}  &90.6           &79.7 &\textbf{92.3} &\textbf{88.4} &85.3 &\textbf{96.1} &\textbf{77.2} &95.3 &\textbf{84.2} &64.2 &\textbf{80.0} &83.0 \\   
    Point2Sequence \cite{liu2019point2sequence}	        &85.2           &82.6           &81.8           &87.5           &77.3           &90.8           &77.1&91.1&86.9&83.9&95.7&70.8&94.6&79.3&58.1&75.2&82.8 \\
    LRC-Net \cite{liu2020lrc} 	                     &85.3           &82.6           &85.2           &87.4           &79.0           &90.7           &80.2&91.3&86.9&84.5&95.5&71.4&93.8&79.4&51.7&75.5&82.6 \\ 
    \textbf{D-Net(sps)}                               &85.7  &83.8  &81.0  &86.9  &78.4          &91.1  &74.7&91.6&87.1&84.4&95.7&71.6&95.1&80.9&58.9&75.3&83.4 \\
    \textbf{D-Net(xyz)}                               &86.0  &\textbf{84.5}  &83.1          &85.2  &78.9           &91.0  &\textbf{81.9}&91.5&86.7&85.0&95.9&72.7&\textbf{95.9}&82.4&58.2&76.6&83.4 \\
    \textbf{D-Net(xyz+nor.)}                                       &\textbf{86.2}  &84.4  &83.7           &\textbf{90.4}  &79.4           &\textbf{91.3}  &80.5&91.8&86.9&85.6&96.0&74.9&95.7&83.5&59.4&78.8&83.2 \\
    \hline
    \end{tabular}}
    \caption{The shape segmentation results (\%) on ShapeNet part segmentation dataset.}
    \label{table:part_segmentaion}
\end{table}
\noindent\textbf{Implementation Details.}
The number of neighboring points $k$ in each local region is set as 20.
And the number of high distinctive points and low distinctive points $N_1$ is set as 320.
For feature extraction, a stacked self-gated CNN at depth four is used, where the output of each self-gated CNN is set as 64, 64, 64, and 128, respectively.  
During the training, we use an ADAM optimizer with an initial learning rate of 0.001, a batch size of 16, and a batch normalization rate of 0.5.
In addition, ReLU is used after each fully connected layer followed by a dropout layer with a drop ratio 0.5 in the fully connected layers.

\begin{table}[htp]
\begin{center}
\scalebox{0.8}{
\begin{tabular}{ccccccc}
\hline
$k$ &5  &10 &15 &20  &25 &30   \\ \hline
Instance acc. (\%) &91.94	&92.54 &92.75	&\textbf{93.15}	&93.07 &92.75  \\ \hline
\end{tabular}}
\end{center}
 \caption{The effect of the number of neighbors $k$ under ModelNet40.}
\label{table:k}
\end{table}

\noindent\textbf{Parameters.}
All the experiments in this section were evaluated on ModelNet40, which contains 40 categories and 12,311 CAD shapes with 9,843 shapes for training and 2,468 shapes for testing, respectively.
For each 3D shape, we adopt the point cloud with 1,024 points \cite{qi2017pointnet,qi2017pointnet++,wang2019dynamic} which are uniformly sampled from the corresponding mesh as input.

In the module of distinctive feature extracting, the number of neighbors $k$ in each local region is an important parameter, which influences the capturing of local structures around each point.
The results of several settings of $k$ are shown in Table \ref{table:k}.
The best instance accuracy $93.15\%$ is reached at $k = 20$, which can better cover the context information inside local regions.
From the results, we can see that capturing local structures plays an important role in enhancing the global representation learning of point clouds.

\begin{table}[h]
    \begin{center}
    \scalebox{0.9}{
    \begin{tabular}{ccccccc}
    \hline
    $N_1$ &384 &320 &256 &192 &128    \\ \hline
    Instance acc. (\%) &92.83 &\textbf{93.15} &92.63 &92.79 &92.34 \\ \hline
    \end{tabular}}
    \end{center}
    \caption{The effect of distinctive point set size $N_1$ under ModelNet40.}
    \label{table:N1}
\end{table}

Finally, we explore the effect of the distinctive point set size $N_1$ under ModelNet40.
$N_1$ is also an important parameter that influences the capturing of the structure information within distinctive point sets.
Due to the variance of point clouds, the frequency of distinctive points and non-distinctive points are likely to be class-dependent or even shape-dependent. To simplify the parameter setting, we set the number of distinctive and non-distinctive points as a same value $N_1$. And in order to further explore the influence of $N_1$, we have conducted parameter comparison in Table \ref{table:N1}. 
As shown in Table \ref{table:N1}, the best install accuracy is achieved at $N_1 = 320$, which better covers the entire training data. 

\begin{figure}[ht]
    \centering
    \includegraphics[width=\textwidth]{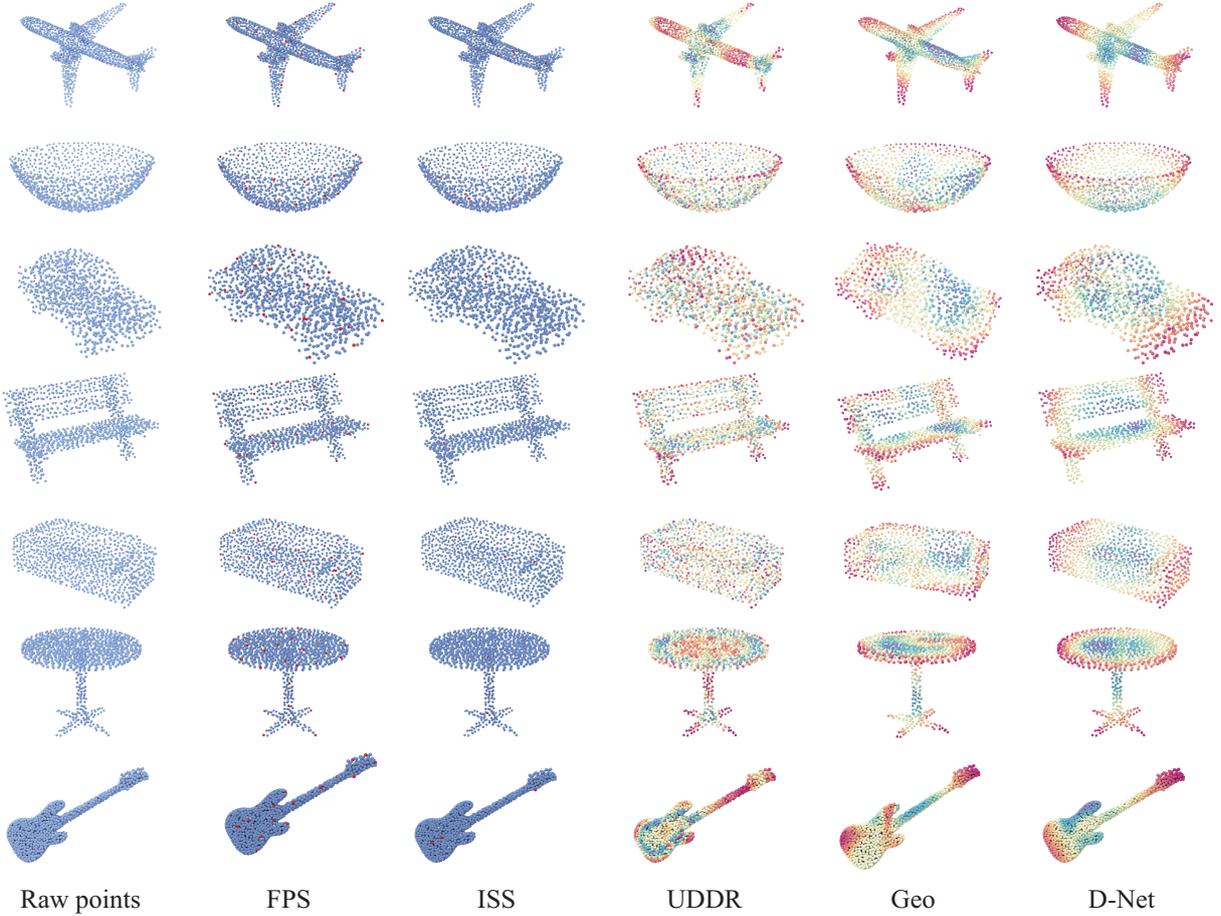}
    \caption{The comparisons with existing methods on distinction detection.}
    \label{fig:detection}
\end{figure}
\noindent\textbf{Shape Classification.}
We evaluate D-Net on ModelNet40 classification benchmark \cite{wu20153d}, which is composed of 9,843 train models and 2,468 test models in 40 classes. 
Same as \cite{qi2017pointnet,qi2017pointnet++}, we sample 1,024 points and normalize them to a unit sphere for each point cloud.

The quantitative comparisons with the state-of-the-art point-based methods are summarized in Table \ref{table:cls}, where D-Net outperforms all the xyz-input methods.
On the ModelNet40 dataset, our approach performs best among the compared methods, reaching $93.2\%$ in instance accuracy.
We further test our method with point normals.
For a fair comparison, we report the results of RS-CNN without voting strategy according to their paper \cite{liu2019relation}.
In addition, we also show the results of D-Net in unsupervised shape classification under the ModelNet40 benchmark.
Following PointNet++ \cite{qi2017pointnet++} and LGAN \cite{achlioptas2017learning}, we integrate feature interpolation for point upsampling and earth mover's distance (EMD) as the training loss to formulate D-Net++.
And we employ multi-class SVM to be the classifier as in \cite{liu2019l2g}.
D-Net also achieves state-of-the-art performances in the unsupervised shape classification application.

To compare the point distinction detection performances, we revisit the public distinction detection methods and correspnding source code. we compare our D-Net with existing methods, including Farthest point sampling (FPS), ISS \cite{zhong2009intrinsic} and UDDR \cite{li2020unsupervised} as shown in Fig. \ref{fig:detection}.
Specifically, FPS is a widely applied algorithm for uniform sampling of point clouds, which selects points according to the point-to-point distance and ignores the point distinction. Here, we select 32 points from each raw point cloud (1024 points). The ISS algorithm is a traditional key point detection method, which calculates the point distinction via the geometry distribution of points. Due to the predefined threshold value, the number of selected key points in ISS algorithm is unstable. And the UDDR is a learning-based distinctive region detection method, which learns the point distinction with a clustering-based nonparametric softmax classifier in an iterative re-clustering manner. Similarly, both UDDR and D-Net predict a distinction score for each point, where the warmth of the point color indicates the magnitude of the distinction score. From the visualization results, we can see that both UDDR and D-Net tend to assign higher distinction scores to edge points or sharp points. And the distribution of the distinction point is largely determined by the design of different strategies, such as the distinction metric and the task optimization. Overall, different from existing methods, our D-Net focuses more on extracting high-frequency information, such as corners or edges, to learn discriminative point cloud representations. And the experimental results have proved the effectiveness of our strategy to explore the point distinction in the feature space. 


\begin{figure}[t]
    \begin{minipage}[t]{0.45\textwidth}
    \centering
    \includegraphics[height=5cm]{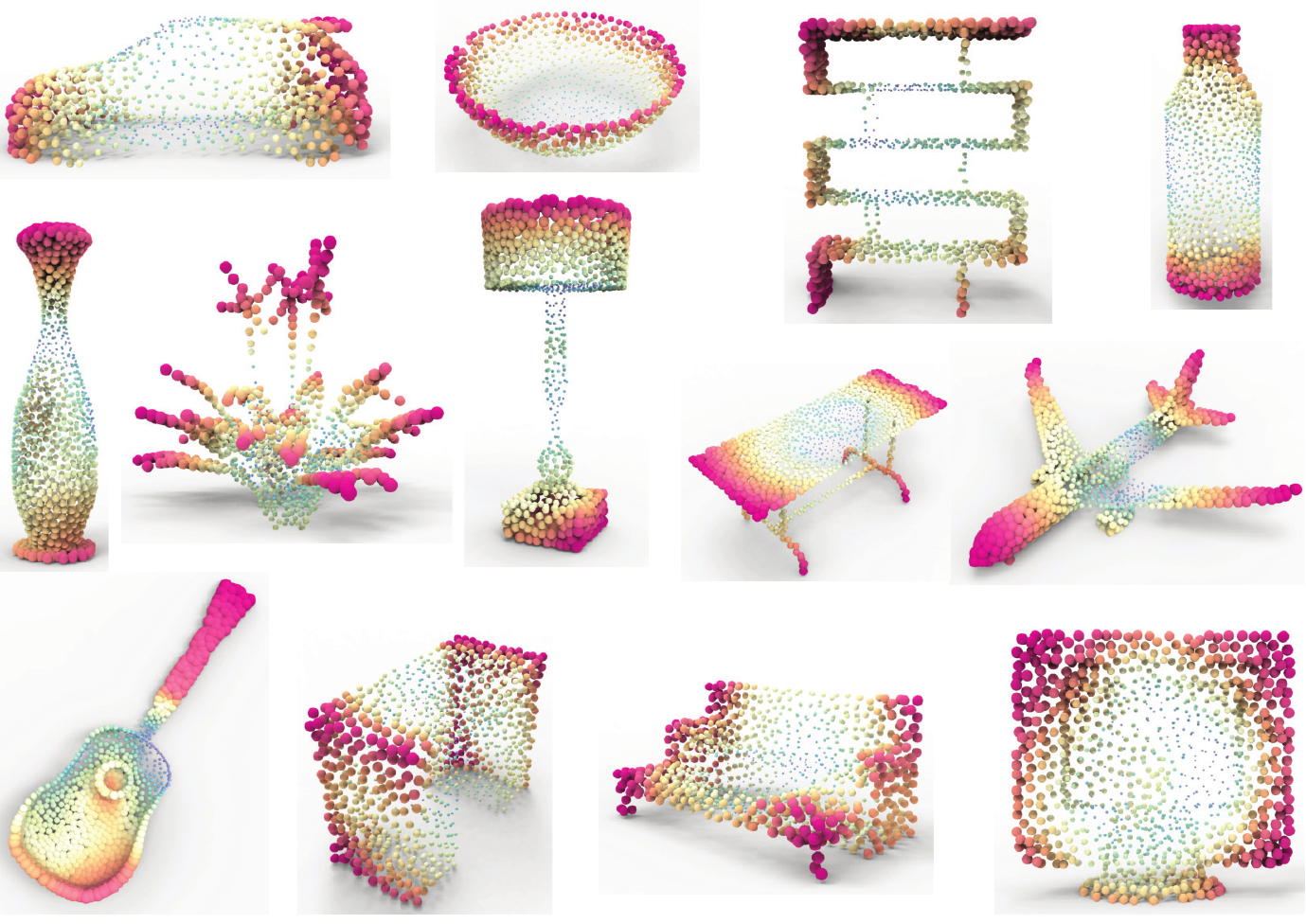}
    \caption{The visualization of point distinction under ModelNet40.}
    \label{fig:dps}
    \end{minipage}
    \begin{minipage}[t]{0.55\textwidth}
    \centering
    \includegraphics[height=5cm]{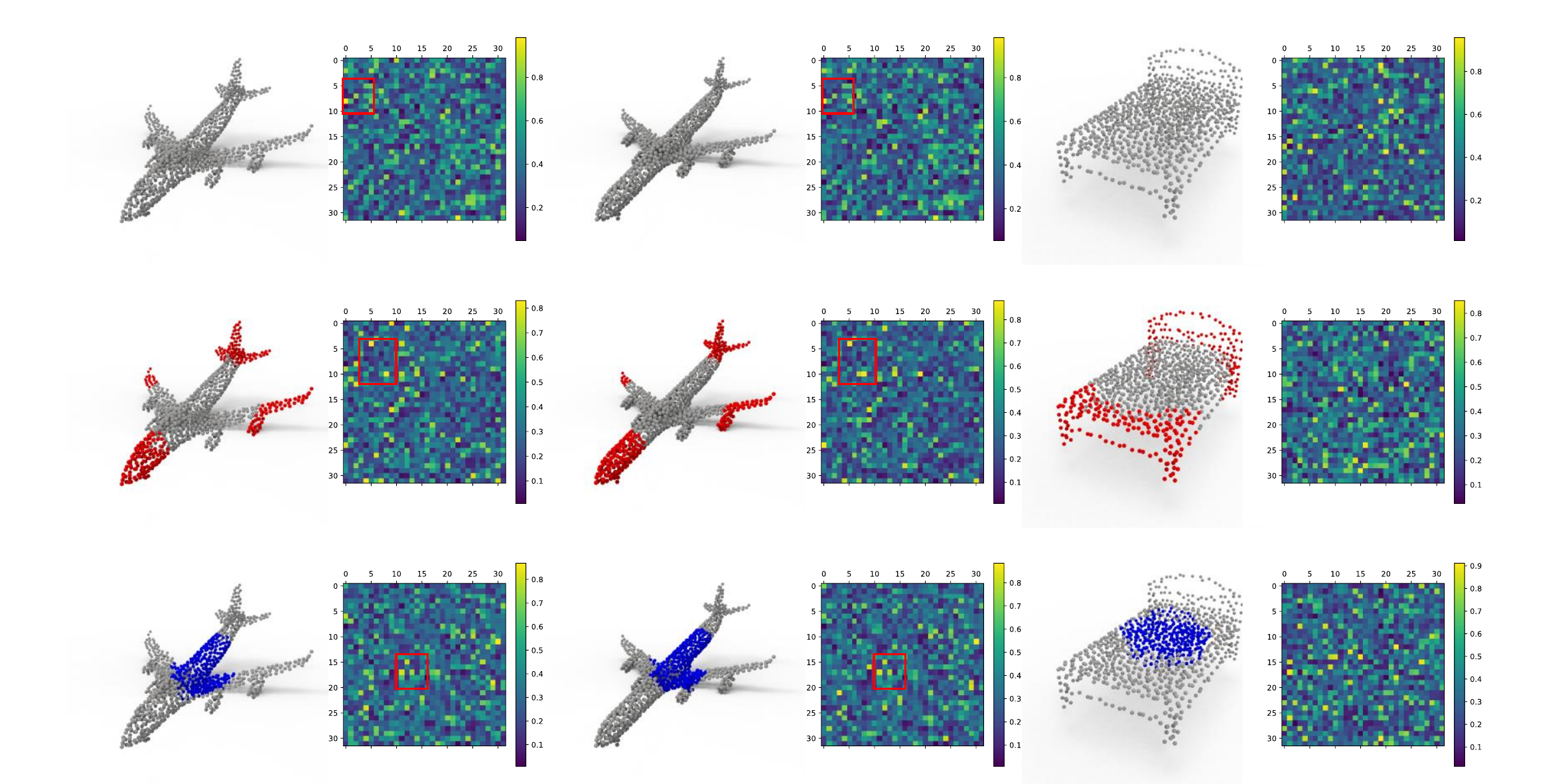}
    \caption{The visualization of feature fusion weights.}
    \label{fig:llfw}
    \end{minipage}
\end{figure}

\begin{figure}[t]
    \centering
    \includegraphics[width=\textwidth]{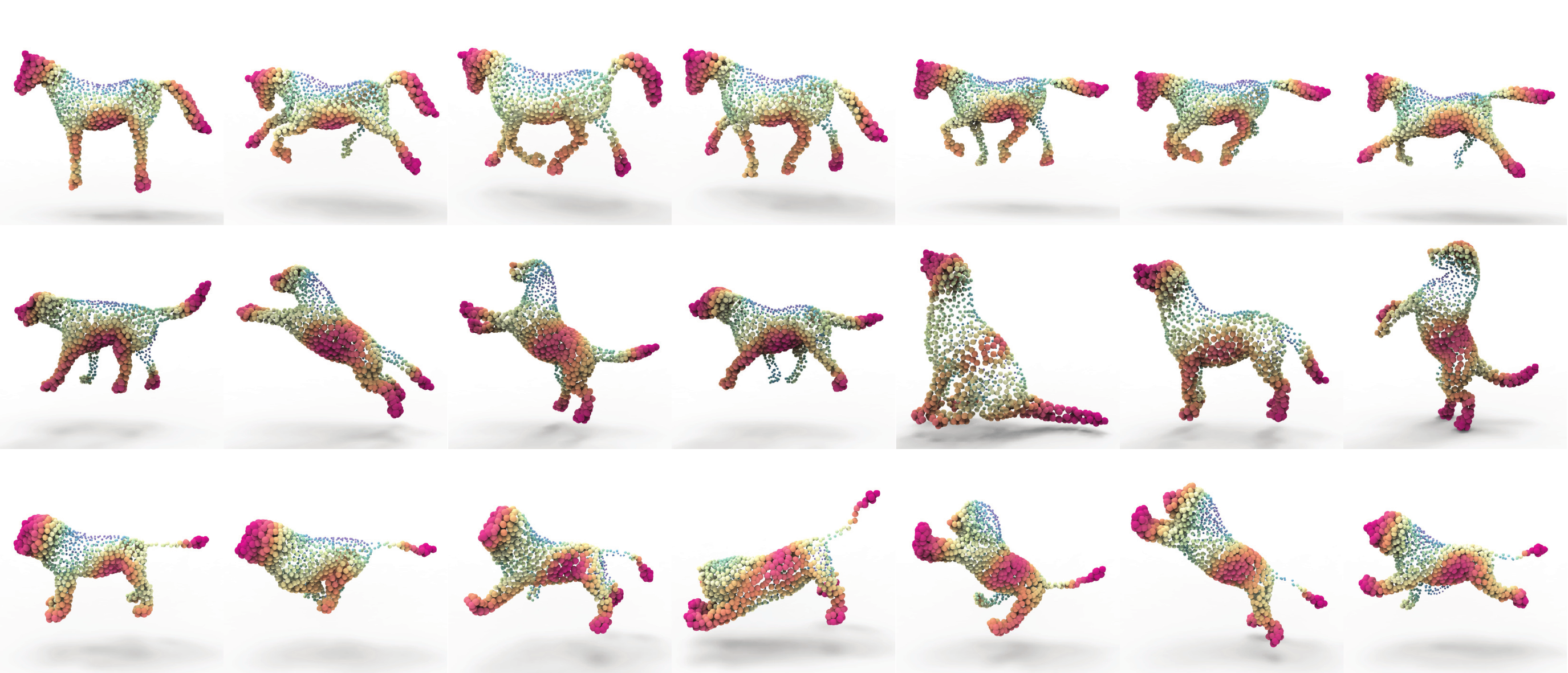}
    \caption{Consistent distinction distribution under ISDB \cite{gal2007pose}.}
    \label{fig:consistent}
\end{figure}

\noindent\textbf{Shape Part Segmentation.}
Part segmentation is a challenging point cloud analysis task, which predicts a semantic part label for each input point.
We evaluate D-Net in shape part segmentation on ShapeNet part benchmark \cite{yi2016scalable}.
This dataset consists of 16,881 models from 16 shape categories and is labeled in 50 part classes in total.
We follow \cite{qi2017pointnet} to split shapes into the training set and test set with 2,048 points for each point cloud.
For comparison, we report the mean instance IoU (Intersection-over-Union) that is averaged across all instance point clouds.

\begin{table}[h]
    \centering
    \begin{tabular}{ccccccccc}
        \hline
        Metric &Geo &FPS &RS &NoSG &Max &Mean &SC &ALL \\ 
        Acc. &92.78 &92.71 &92.79 &91.08 &92.34 &92.63 &92.63 &\textbf{93.15}\\ \hline
    \end{tabular}
    \caption{The ablation study under ModelNet40.}
    \label{table:ablation}
\end{table}

Table \ref{table:part_segmentaion} shows the quantitative comparisons with the state-of-the-art approaches, where D-Net achieves the best performance with an instance mIoU of 86.2\%.
We resort to FPS to select points from the raw point cloud, where Farthest Point Sampling (FPS) can better capture the underlying shape for point clouds in part segmentation.
The results of different settings are shown in  Table \ref{table:part_segmentaion}, including input with xyz only (xyz), with xyz and normal (xyz,nor.), and using self-attentive point searching (sps).
To qualitatively show the performance of D-Net, we also visualize some examples of prediction results, where our results are high consistency with the ground-truth as shown in Fig. \ref{fig:sem}.

\noindent\textbf{Ablation Study.}
To quantitatively evaluate the effect of proposed modules, we show the performances of D-Net under three settings: without self-attentive point searching module (NoSPS), without self-gate in the convolution module (NoSG), and without learnable feature fusion module (NoLFF).
Specifically, in NoSPS, we utilize FPS (FPS) and random sampling (RS) to replace the self-attentive point searching.
And in the NoLFF, we replace the learnable gate with max-pooling (Max), mean pooling (Mean), or simple concatenation (SC).
In addition, we also show the result of D-Net with all proposed modules (ALL).
As shown in Table \ref{table:ablation}, we report the results of D-Net with the above settings.
Each module effectively learns the discriminative point cloud representation, which distinguishes a shape from other classes.
And to explore the geodesic distance in searching the distinctive points, we also show the results of extracting local features with geodesic distances (Geo). 
To calculate the geodesic distance between points, we directly apply the Isomap \cite{tenenbaum2000global} with the Python implementation for approximating the geodesic distance between points.
Thus, we adopt point geodesic distances to search neighbor points for extracting point features from local regions.
Due to the limitation of approximation accuracy and large computational complexity, the geodesic distance is not performing well in the point cloud classification. In addition, we also report the distinction detection results under the usage of point geodesic distance.

To explore the effect of the distinctive point set, we adjust the used distinctive point set as shown in Table \ref{table:dis}.
Specifically, we report several results, including raw point cloud only ($P_R$), raw point cloud and high distinctive point set ($P_R + P_H$), high distinctive point set only ($P_H$), high distinctive point set, low distinctive point set ($P_H + P_L$), and all distinctive point sets ($P_R+P_H+P_L, ALL$).

\begin{figure}[t]
    \centering
    \includegraphics[width=\textwidth]{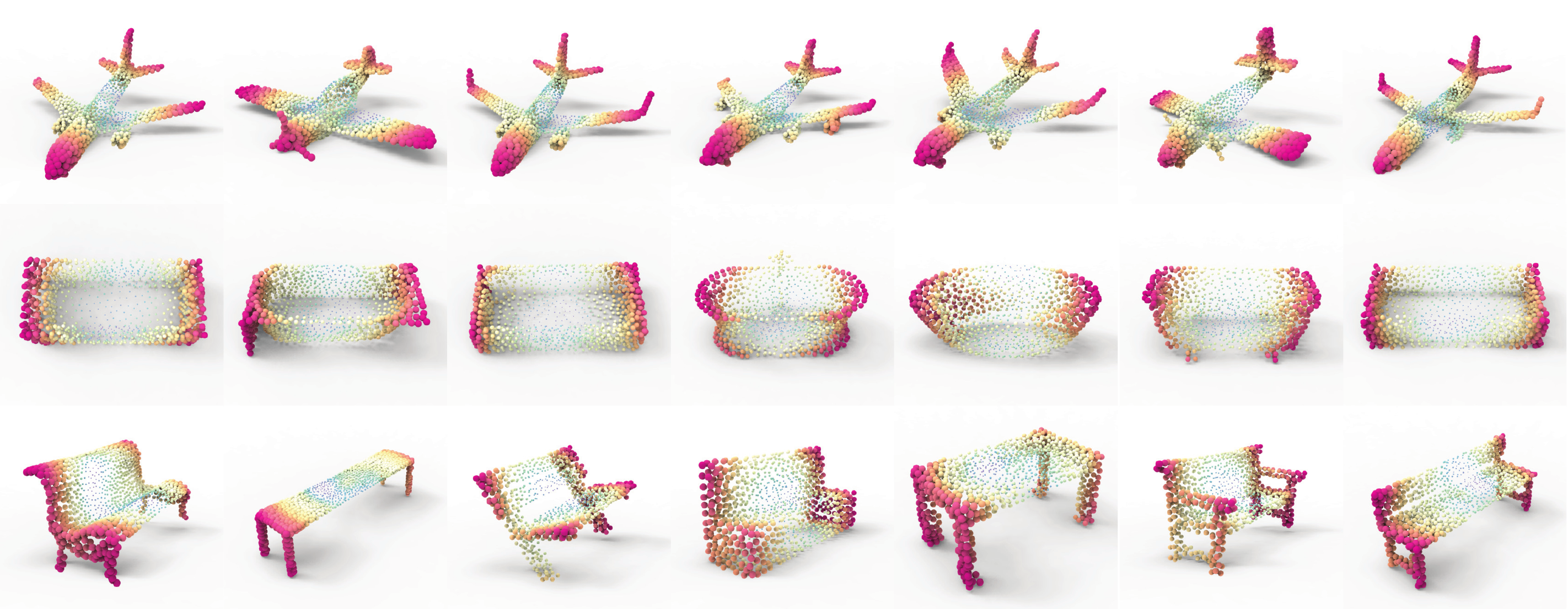}
    \caption{Shape consistency under ModelNet40.}
    \label{fig:MN_consistent}
\end{figure}

\begin{table}[h]
\begin{center}
\begin{tabular}{cccccc}
\hline
Metric &$P_R$ &$P_R + P_H$ &$P_H$ &$P_H + P_L$ &$ALL$    \\
Instance acc. (\%) &92.34 &92.54 &90.7 &92.1 &\textbf{93.15}	\\ \hline
\end{tabular}
\end{center}
\caption{The effect of used point sets under ModelNet40.}
\label{table:dis}
\end{table}
\noindent\textbf{Visualization Analysis.}
In D-Net, there are some important visualization results that should be shown, including distinctive point clouds and learnable feature fusion weights.
In Fig. \ref{fig:dps}, we show some examples of the distinctive point clouds, where the self-attentive point searching captures a distinction score for each point. 
In addition, we also show the learnable feature fusion weights in Fig. \ref{fig:llfw}, where the first row is the raw point cloud and corresponding fusion weights.
In particular, we resize each 1024-dimensional weight vector into $32\times32$ matrix for visualization.
Similar to the first row, the second row represents the high distinctive point sets, and the third row represents low distinctive point sets.
And we have marked several similar areas with red boxes in the weight matrixes.
For two airplanes from the same class in ModelNet40, the weight matrix shows the consistency between the same class and the difference between different classes.

\begin{figure}[t]
    \centering
    \includegraphics[width=12cm]{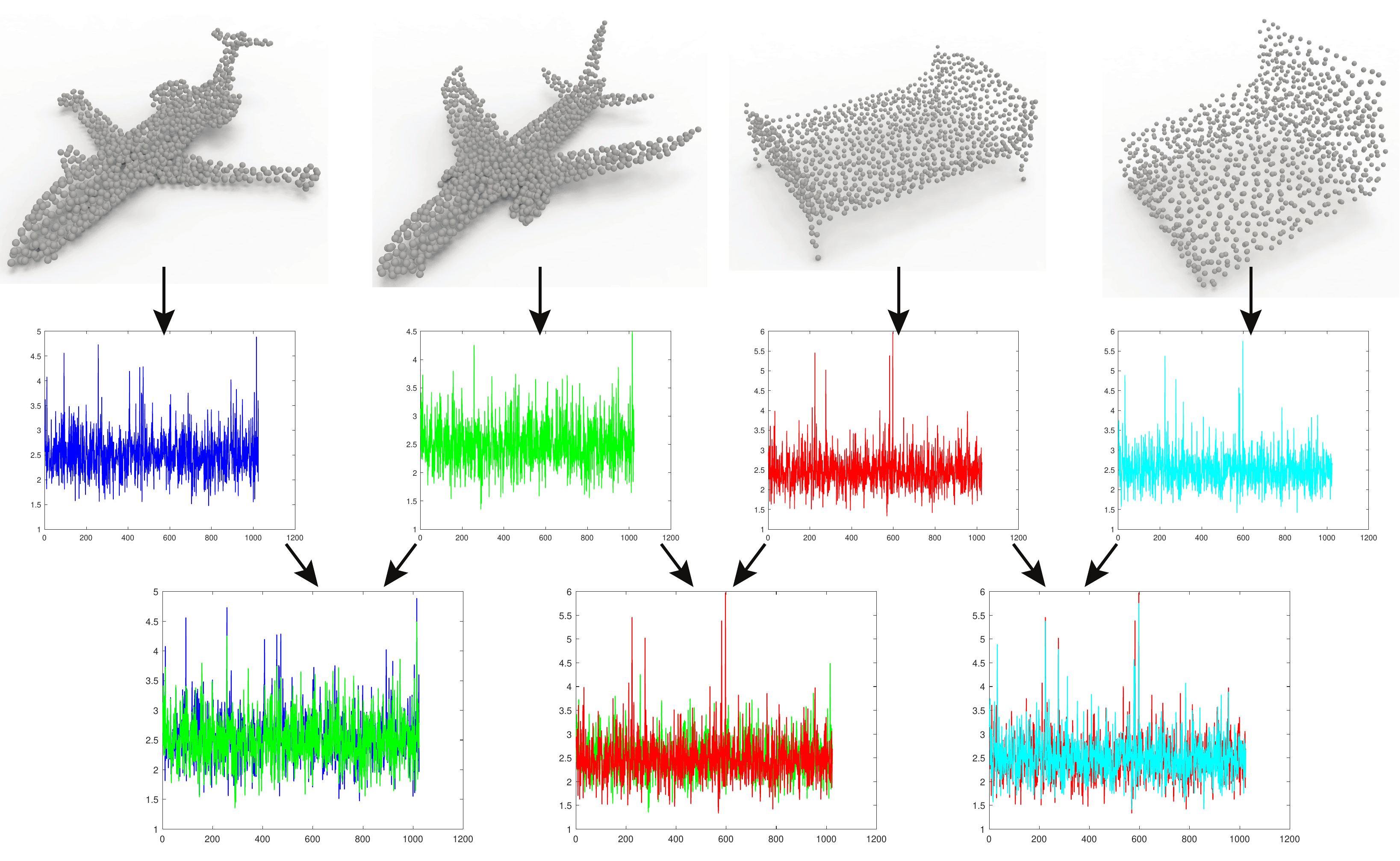}
    \caption{The visualization of feature consistency. The first row contains four 3D point clouds, including two airplanes and two beds.
    And the second row indicates the corresponding learned feature distribution of corresponding point clouds, where each feature is extracted by D-Net under ModelNet40.
    To show the consistency between the learned features, we show the overlapping of different feature distributions in the last row, where different features are mixed as shown by arrows.}
    \label{fig:consistency}
\end{figure}
To further evaluate our D-Net in terms of effectively learning distinctive points, we show the visualization of some distinctive point clouds from ISDB \cite{gal2007pose}.
ISDB is a database of different articulated models of animals and humans containing about 104 models.
We sample 1,024 points from each mesh shape with Poisson Disk Sampling \cite{corsini2012efficient} and directly process these points with a network trained under ModelNet40.
In Fig. \ref{fig:consistent}, we visualize the distinctive point clouds of horses, dogs, and lions.
The visualization results suggest that our method can learn shapes with different poses with high consistency within the same class, such as horse and dog.
In addition, the distinction distribution of shapes under ModelNet40 also shows high consistency, as shown in Fig. \ref{fig:MN_consistent}.

In D-Net, we propose a self-attentive point searching module to capture an important score for each point.
As shown in the experiments, the distinction distribution of points tends to be consistent within a class.
To further reveal the consistency of the learned features, we draw the feature distribution as shown in Fig. \ref{fig:consistency}.
The horizontal axis represents the feature channel (1,024-dimensional), and the vertical axis represents the feature value of each channel, such as airplanes and beds.
From the visualization, we can know that the learned features from the same class have a close feature distribution and high feature consistency.
Otherwise, the learned features from different classes have low feature consistency.

\begin{table}[htp]
\begin{center}
\begin{tabular}{lccc}
\hline
Method &Model size (MB) &Time (ms) & Accuracy (\%)   \\ \hline
PointNet (vanilla) \cite{qi2017pointnet}       &9.4   &6.8 &87.1	\\
PointNet \cite{qi2017pointnet}                  &40 &16.6  &89.2\\
PointNet++ (SSG) \cite{qi2017pointnet++}       &8.7 &82.4 &- \\
PointNet++ (MSG) \cite{qi2017pointnet++}          &12 &163.2 &90.7 \\
LRC-Net \cite{liu2020lrc}             &18 &115.8 &93.1  \\ 
DGCNN \cite{wang2019dynamic}          &21 &27.2 &92.9 \\
PointCNN \cite{li2018pointcnn}       &94 &117.0 &92.3   \\
D-Net (ours)                        &28.8 &85.8 &93.2 \\ \hline
\end{tabular}
\end{center}
\caption{Complexity, forward time, and accuracy of different models under ModelNet40.}
\label{table:com}
\end{table}
\section{Model Complexity}
In addition, to show the network complexity of D-Net intuitively, we make statistics of model size and computational cost of some point cloud based methods.
We follow PointNet++ to evaluate the time and space cost of several point cloud based methods as shown in Table \ref{table:com}. 
We record forward time under the same conditions with a batch size 8 using TensorFlow 1.0 using a single GTX 1080 Ti. 
Table \ref{table:com} shows D-Net can achieve a trade-off between the model complexity (number of parameters) and computational complexity (forward pass time).
For a fair comparison, we report the model complexity of PointNet++ \cite{qi2017pointnet++} and DGCNN \cite{wang2019dynamic}, which are the basis of our D-Net.
In addition, we also show the complexity of some recent works including PointCNN \cite{li2018pointcnn} and LRC-Net \cite{liu2020lrc}.

\section{Conclusion}
In this paper, D-Net (Distinctive Network) is proposed to capture the distinction of points for global feature learning in point cloud analysis tasks.
The core of D-Net is point importance capturing, which successfully learns a distinction score for each point with self-attentive point searching.
With multiple distinctive point sets, the proposed stacked self-gated convolution module can effectively extract distinctive set features.
In addition, our learnable feature fusion mechanism effectively aggregates distinctive point set features into a global point cloud representation.
Our superior over other methods in experiments demonstrates that D-Net is effective in the learning of distinction for points, which can be further leveraged for better point cloud analysis.

\section*{Acknowledgements}

The corresponding author is Yu-Shen Liu. This work was supported by National Key R\&D Program of China (2022YFC3800600), the National Natural Science Foundation of China (62272263, 62072268), and in part by Tsinghua-Kuaishou Institute of Future Media Data.

\bibliography{mybibfile}

\end{document}